\documentclass{article}

\usepackage[numbers, super]{natbib}
\usepackage{threeparttable}
\usepackage{booktabs}

\makeatletter
\renewcommand\@biblabel[1]{#1.}
\makeatother
\usepackage{caption}
\usepackage{float}
\usepackage{graphicx}
\usepackage[utf8]{inputenc} 
\usepackage[T1]{fontenc}    
\usepackage{hyperref}       
\usepackage{url}            
\usepackage{booktabs}       
\usepackage{amsfonts}       
\usepackage{nicefrac}       
\usepackage{microtype}      
\usepackage{xcolor}         
\usepackage{multicol}

\usepackage[final]{neurips_2023}

\title{AlgoRxplorers | Precision in Mutation \\ Enhancing Drug Design with Advanced Protein Stability Prediction Tools\thanks{This paper was authored in April 2024 as a class project for Data and Visual Analytics (CSE 6242) at Georgia Tech and was published on arXiv in January 2025.}}

\author{%
  Karishma Thakrar \\
  Georgia Institute of Technology \\
  \texttt{karishma.thakrar@gatech.edu} \\
  \AND
  Jiangqin Ma \\
  Georgia Institute of Technology \\
  \texttt{jma416@gatech.edu} \\
  \And
  Max Diamond \\
  Georgia Institute of Technology \\
  \texttt{mdiamond32@gatech.edu} \\
      \And
  Akash Patel \\
  Georgia Institute of Technology
  \\ \texttt{apatel969@gatech.edu} \\
}

\begin{document}

\maketitle


\begin{abstract}

Predicting the impact of single-point amino acid mutations on protein stability is essential for understanding disease mechanisms and advancing drug development. Protein stability, quantified by changes in Gibbs free energy ($\Delta\Delta G$), is influenced by these mutations. However, the scarcity of data and the complexity of model interpretation pose challenges in accurately predicting stability changes. This study proposes the application of deep neural networks, leveraging transfer learning and fusing complementary information from different models, to create a feature-rich representation of the protein stability landscape. We developed four models, with our third model, ThermoMPNN+, demonstrating the best performance in predicting $\Delta\Delta G$ values. This approach, which integrates diverse feature sets and embeddings through latent transfusion techniques, aims to refine $\Delta\Delta G$ predictions and contribute to a deeper understanding of protein dynamics, potentially leading to advancements in disease research and drug discovery.

\end{abstract}

\section{Introduction}

Many drugs target proteins to modulate their activity such that they may bind more efficiently to their targets and lead to more effective treatments \cite{Baek2022}. A protein's activity is typically impacted by alterations in its sequence, leading to a significant change in its structure and function, thereby influencing protein stability. When mutations cause the protein to become unstable, it often leads to a range of diseases and cancers, underscoring the importance of maintaining protein structure integrity for cellular health. Meanwhile, mutations that enhance protein stability could lead to the development of more effective drugs, offering new avenues for treatment. In 2022 alone, pharmaceutical companies spent nearly \$244 billion on R\&D, underscoring the importance of a more efficient drug discovery and development process \cite{WinNT}. We aim to reduce the time taken to identify viable new drug candidates by assessing the impact of protein point mutations on stability. This will be done by analyzing the change in Gibbs free energy ($\Delta$$\Delta$G) between naturally occurring, or wild-type, proteins and their mutated versions, which is a common way to assess stability changes. Our goal is to create an algorithm that predicts changes in Gibbs free energy caused by single-point amino acid mutations, effectively forecasting how specific mutations can alter protein stability.

\section{Current Practice}

Our project draws upon a diverse range of recent scientific research to explore innovative protein stability prediction techniques.

Cao \cite{cao2019}, Heyrati \cite{Heyrati2023}, Kuhlman \cite{Kuhlman2019}, Dieckhaus \cite{Dieckhaus2024}, M. Baek \cite{Baek2022}, Chandra \cite{Chandra2023} and Wei \cite{Wei2019} introduce neural network models for predicting protein characteristics, ranging from stability change to bioactivity and even protein structure. Cao's \cite{cao2019} DeepDDG model, while directly related to our project, is hard to implement given the code isn't open to the public but its DeepDDG server \cite{deepddgserver} could be used for benchmarking purposes. While Heyrati's \cite{Heyrati2023} study with a sophisticated Siamese neural network allowed us to learn about feature extraction and similarity learning which our algorithms will also be leveraging, their focus on bioactivity classification offered little overlap with ours. Although it didn't examine proteins beyond their structure, we found Kuhlman's \cite{Kuhlman2019} combined use of neural networks along with traditional machine learning methods like SVM on embeddings interesting since we'll also be implementing these methods.

Although M. Baek's \cite{Baek2022} work utilizes amino acids to predict proteins structure, which is valuable for our research, it failed to generalize across diverse protein families. Chandra's \cite{Chandra2023} use of a Transformer and encoded features to predict protein structure was valuable for our research although the training dataset was limited. Wei's \cite{Wei2019}  paper offered a comprehensive overview of applying machine learning techniques to materials science, making otherwise time-consuming tasks more efficient, which resonates with our work. However, the models offered little interpretability which is crucial for trust and adoption.

J. Baek \cite{Baek2021} introduces the Graph Multiset Transformer (GMT) to predict molecular properties. The paper is valuable for us in learning more about the intricate relationships among amino acids, the building blocks of proteins, despite the limited training dataset. M. Baek \cite{Baek2022}, Maziarka \cite{Maziarka2024}, Dieckhaus \cite{Dieckhaus2024} and Wang \cite{Wang2022} extend the research body to GNNs, which help compress complex protein features while preserving key structural interactions. 

Maziarka \cite{Maziarka2024} introduces the Molecule Attention Transformer designed for small molecules which would require significant modifications for us given the complexity of protein structures; the paper was still useful to learn of the various Transformer architectures in the domain. While Wang\cite{Wang2022} offered a small training dataset, the GNNs employed of protein structures for property prediction offered a useful perspective. Dieckhaus \cite{Dieckhaus2024} introduces ThermoMPNN, a model combining GNNs and transfer learning, to predict protein stability changes. ThermoMPNN is relevant for enhancing prediction accuracy by learning from extensive protein behaviors, but the model could be simplified for computational efficiency. Baselious \cite{Baselious2023} and Rives \cite{Rives2020} similarly leverage transfer learning to predict protein structure and properties, respectively.

Baselious \cite{Baselious2023} discusses the use of an advanced protein structure prediction algorithm for proteins that play a major role in epigenetic regulation, developing our understanding of the impact of mutations on protein structure and function despite have inconsistencies in accurately predicting the dynamic regions of proteins. Rives \cite{Rives2020} introduces ESM, a pretrained protein transformer model, which clusters similar proteins. The features generated by ESM, while underfitting the data slightly, will assist our model's ability to develop a rich understanding of proteins and can pair with a simplified ThermoMPNN architecture to generate strong protein stability change predictions.

By leveraging insights across these studies, we will improve upon existing limitations by focusing on building a precise and understandable model.

\section{Methodology}

\hspace{0.25cm} Our novel approach combines advanced deep learning techniques and visualization to improve protein thermostability prediction and streamline the drug discovery process. The modeling component utilizes latent transfusion, fusing latent features or embeddings learned by multiple deep learning models. We leverage FireProtDB, a database providing annotated data on protein stability changes due to mutations and 3D protein structures from experimental findings.

The visualization component allows researchers to explore protein structures throughout the mutation process, along with key metrics related to protein stability. By integrating modeling and visualization, we aim to develop a robust and accurate predictor that quantifies the effects of amino acid mutations on protein stability. Our approach aims to outperform existing methods even with limited training examples, significantly reducing the time and cost associated with drug discovery. This innovative solution tackles the complex problem of protein thermostability prediction, offering a powerful tool to accelerate and optimize the development of novel therapeutics.

\subsection{Data Preprocessing \& EDA}
\hspace{0.25cm} Before modeling, we carefully examined the FireProtDB dataset, which consists of unique proteins, each with several listed mutations. The dataset contains nearly 3,100 mutations. After adjusting for duplicates, accounting for 0.52\% of the training data and 1\% of the validation data, there were a total of 2,645 mutations for training and 398 mutations for validation. 

Through exploratory data analysis, we constructed a confusion matrix comparing wild-type and mutation cases, revealing that the amino acids valine (V) and leucine (L) in the wild-type were frequently replaced by alanine (A) due to mutation. We applied the k-means clustering algorithm to both the ThermoMPNN and XGBoost embedding datasets, with PCA for dimensionality reduction. Using twelve clusters, all data points were perfectly allocated to their appropriate clusters, indicating distinct groups within the dataset. Additionally, we observed that the Immunoglobulin G-binding protein, glycine, was one of the most prominent proteins in the dataset, with a count of 600 mutations. The exploration and preprocessing of the dataset provided valuable insights into the data's characteristics and ensured a solid foundation for building accurate and reliable predictive models.

\subsection{Model Development}

\hspace{0.25cm} Our research builds upon ThermoMPNN, a cutting-edge graph neural network model that utilizes transfer learning to extract structural features and sequence embeddings from the ProteinMPNN model. By taking 3D protein maps as input, ThermoMPNN captures the spatial proximity of protein components and the wild-type protein structure. The model utilizes encoder-decoder layers to process the protein's structural information, effectively creating a residue "social network." Through the use of a light attention mechanism and a multilayer perceptron, ThermoMPNN refines the input data to predict stability changes resulting from mutations.

We also incorporated insights from Chris Deo-tte's Novozymes research, which used Meta's Evolutionary Scale Modeling (ESM), a pretrained protein transformer model, to include more protein features in our models. We made significant adjustments to adapt Deotte's models to our dataset such as correcting the backbone atoms used, requiring us to further develop domain knowledge and our understanding of computational chemistry for protein stability prediction \cite{Deotte2023}.

To augment our model's performance, we explored approaches that integrate feature sets and embeddings from deep thermostability prediction models, such as ThermoMPNN and ESM. By combining these embeddings using element-wise multiplication and concatenation, we create a comprehensive representation of proteins, capturing the specific characteristics of each protein-mutation pair. Element-wise multiplication captures interactions between corresponding features from different embeddings, while concatenation enables learning from a wider range of features. However, the effectiveness of integrating diverse embeddings depends on their ability to capture the complexity of the relevant data, and the increased feature space may not always improve performance. Acknowledging this uncertainty, we adopted an experimental approach, comparing the performance of four different models with unique embedding integration strategies. This iterative process allows us to refine our methods, identify the most promising approach, and develop robust models for predicting protein stability changes. 

Our first model is the original ThermoMPNN model trained on data with duplicates removed, providing a solid foundation but potentially limiting in its ability to capture complex interactions between protein features. The second model concatenates ESM embeddings and ThermoMPNN embeddings after applying light attention to predict $\Delta\Delta G$. By extracting the ThermoMPNN embeddings after the light attention layer, we leverage the layer's ability to refine and focus the embeddings on the most relevant information for predicting $\Delta\Delta G$. While Model 2 increases the feature space and allows learning from both structural and sequence information, it may not fully capture interactions between them.

The third model, ThermoMPNN+, combines structural features and sequence embeddings from the ProteinMPNN model with ESM embeddings through element-wise multiplication. This approach increases the feature space and captures interactions between corresponding features. The combined embeddings are further processed by a light attention layer, refining and focusing the embeddings on the most relevant information for predicting our target variable, followed by an MLP, modeling complex non-linear relationships, similiar to the original ThermoMPNN model. Despite potentially overemphasizing certain features or interactions, ThermoMPNN+ demonstrates the best performance among the four models, with the highest validation R², lowest MSE, and lowest RMSE. These results suggest that integrating diverse feature sets and embeddings through latent transfusion enhances the model's ability to accurately predict protein stability changes. Figure 2 provides a closer look at this model's architecture.

\begin{figure}[h]
    \centering
    \caption{\textbf{ThermoMPNN+ Architecture}}
    \includegraphics[scale=0.375]{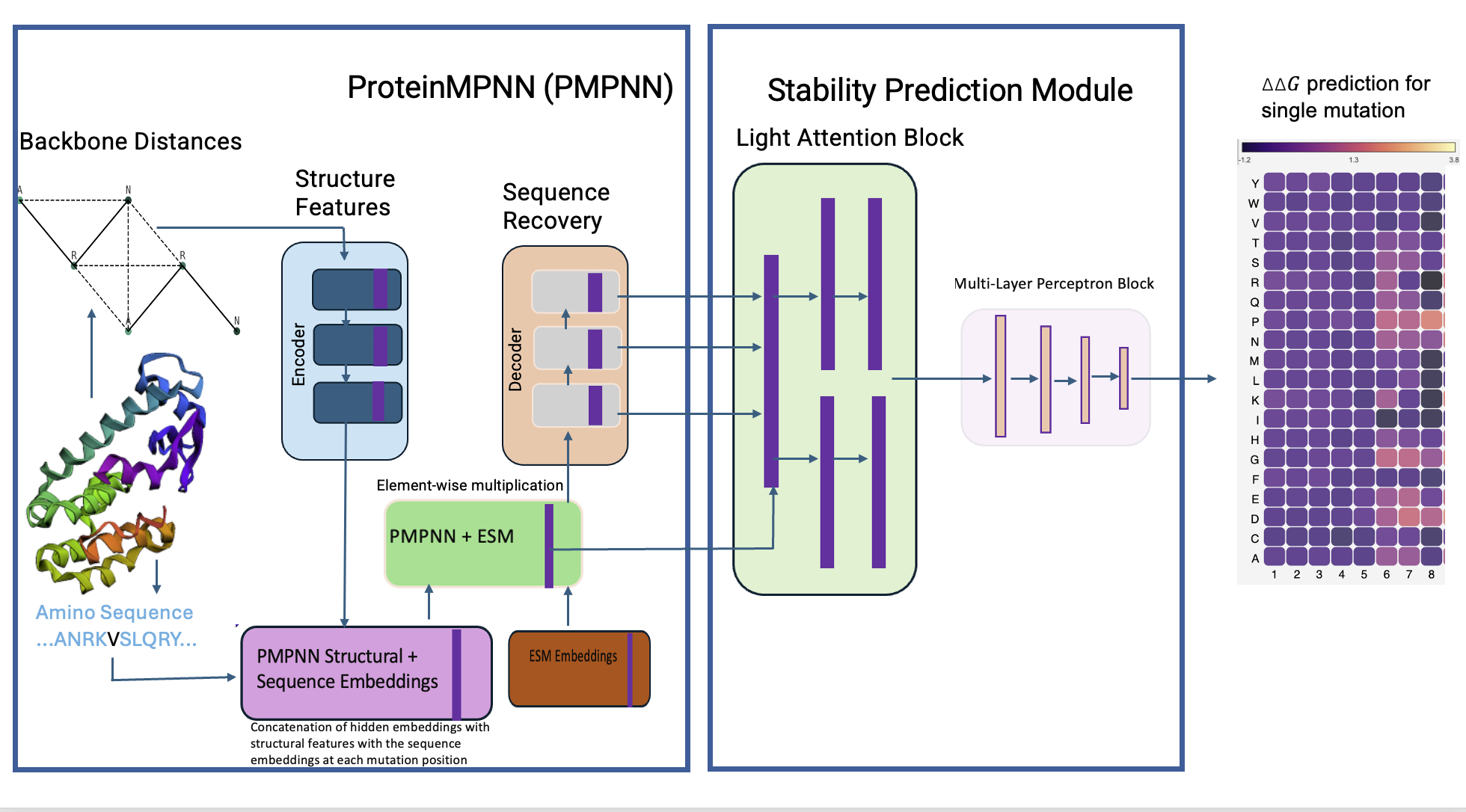}
\end{figure}

The fourth model, heavily influenced by Deotte's work, incorporates a wide range of protein features, including molecular weight, angles among backbone atoms (whereas original ThermoMPNN embeddings were created using 3-D coordinates of backbone atoms), pooled and local ESM embeddings data around mutations, and domain-specific knowledge such as BLOSUM and DeMaSK substitution matrices \cite{Deotte2023}. These matrices quantify the impact of specific amino acid changes on protein stability. The model uses this information to generate trained embeddings, which are concatenated with trained ThermoMPNN embeddings from after the light attention layer to develop a $\Delta\Delta G$ predictor. While this approach leverages a diverse set of features and domain knowledge, it introduces complexity and computational overhead to the model. See Table 1 for a summary of inputs for each model.


\begin{table}[!ht]
    \centering
    \small 
    \caption{\textbf{Models Overview}}
    \begin{tabular}{cp{0.5\textwidth}}
        \toprule
        \textbf{Model} & \textbf{Inputs} \\
        \midrule
        Model 1 & Original ThermoMPNN model inputs: FireProtDB data with duplicates removed \\
        Model 2 & ESM embeddings and trained ThermoMPNN embeddings \\
        Model 3 & ThermoMPNN inputs (structural features and sequence embeddings from ProteinMPNN) combined with ESM embeddings \\
        Model 4 & Trained embeddings generated using domain-specific knowledge combined with ThermoMPNN trained embeddings \\
        \bottomrule
    \end{tabular}
\end{table}

We evaluated performance consistent with the original ThermoMPNN research, using the epoch with the highest Spearman correlation, which measures the monotonic relationship between predicted and actual $\Delta\Delta G$ values, making it less sensitive to outliers. This ensures that the model's predictions align well with the true $\Delta\Delta G$ values, crucial for identifying promising mutations for enhancing protein stability.




\subsection{UI Development}

\hspace{0.25cm} The ThermonMPNN+ modeling is done to enhance predictions of protein thermostability chan-ges, and is supplemented with an innovative web app. Our end product, built with Python, HTML, and JavaScript, complements our innovative approach by providing an interactive 3D visualization of proteins and dashboards for real-time exploration of mutation effects. Designed for researchers and pharmaceutical companies, this tool offers rapid and precise predictions of how mutations impact protein stability, distinguishing our project through its unique combination of interactive visuals and predictive capabilities.

We utilize AlphaFold2, a state-of-the-art artificial intelligence program developed by DeepMind \cite{Jumper2021}, to predicts the 3D structure of proteins with high accuracy based on their amino acid sequence. The web app allows users to visualize and interact with a protein's 3D structures, both wild-type and mutated. An example of the wild-type and mutated 2LZM and 1APS proteins are shown below in Figures 2 and 3, respectively. 

\begin{figure}[!ht]
    \centering
    \caption{\textbf{Interactive 3D protein structures of 2LZM}}
    \includegraphics[scale=0.35]{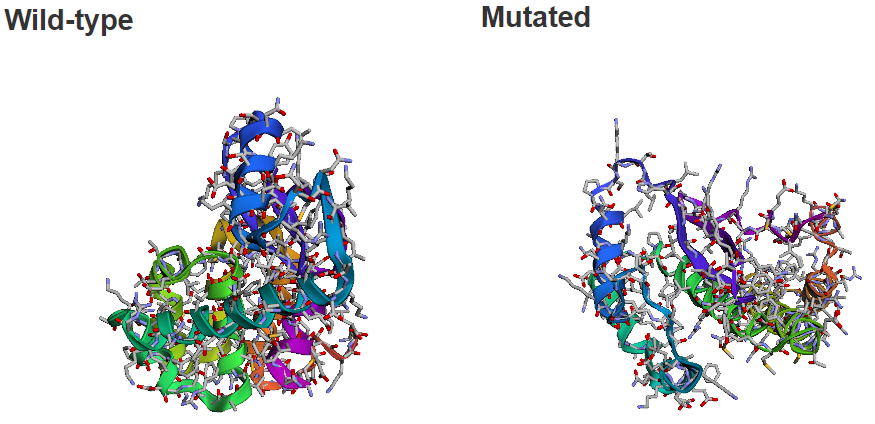}
    \includegraphics[scale=0.8]{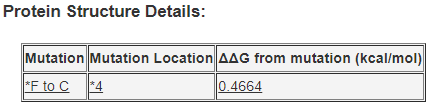}
\end{figure}

\begin{figure}[!ht]
    \centering
    \caption{\textbf{Interactive 3D protein structures of 1APS}}
    \includegraphics[scale=0.35]{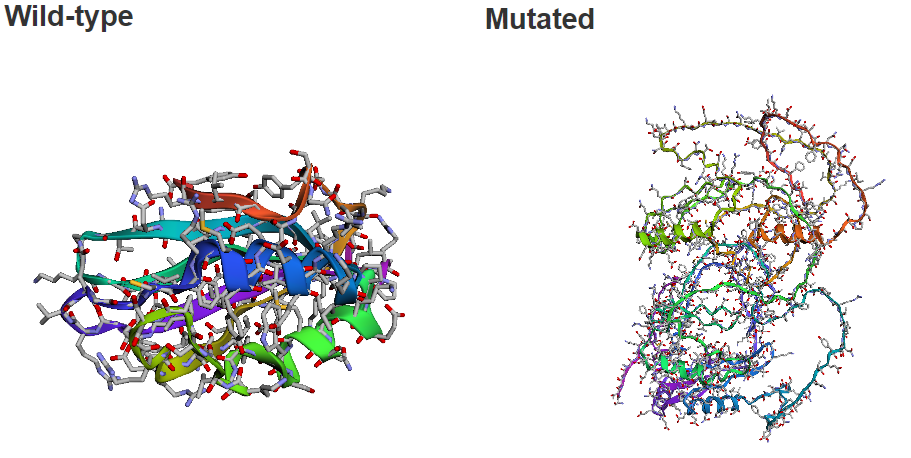}
    \includegraphics[scale=0.8]{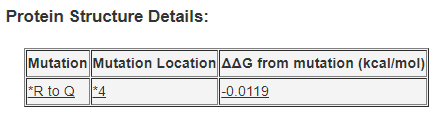}
\end{figure}

The protein structure is displayed using a spectrum color scheme, where the sequence is colored in a gradient transitioning from blue at the N-terminus to red at the C-terminus, highlighting the orientation and folding of the protein chain. For instance, when predicting the structure of protein 2LZM (PDB ID: 2LZM, in Figure 3), minimal deviations were observed from the wild-type structure post-mutation. Conversely, protein 1APS (PDB ID: 1APS, in Figure 4) exhibits substantial structural changes upon similar single-site mutations. These observations underscore the multifaceted nature of protein responses to mutations, influenced by the specific mutation site, amino acid substitution, and the overall protein architecture. It is noteworthy that, while $\Delta$$\Delta$G values provide insights into stability changes, they do not consistently predict the magnitude of structural changes, affirming that minor alterations can significantly destabilize a protein, and substantial structural shifts do not necessarily equate to drastic stability impacts.




Users are also able to explore the change in structure and Gibbs free energy due to single point mutations in real-time. Users can select from a wide array of proteins and mutations to find the best combinations based on their particular use cases. Figure 4, below, illustrates the interactive matrix summarizing the predicted $\Delta\Delta G$ due to a substitution of the amino acid at position i with the amino acid in row j.

\begin{figure}[!ht]
    \centering
    \caption{\textbf{Predicted $\Delta\Delta G$ interactive matrix by mutation type and position}}
    \includegraphics[scale=0.25]{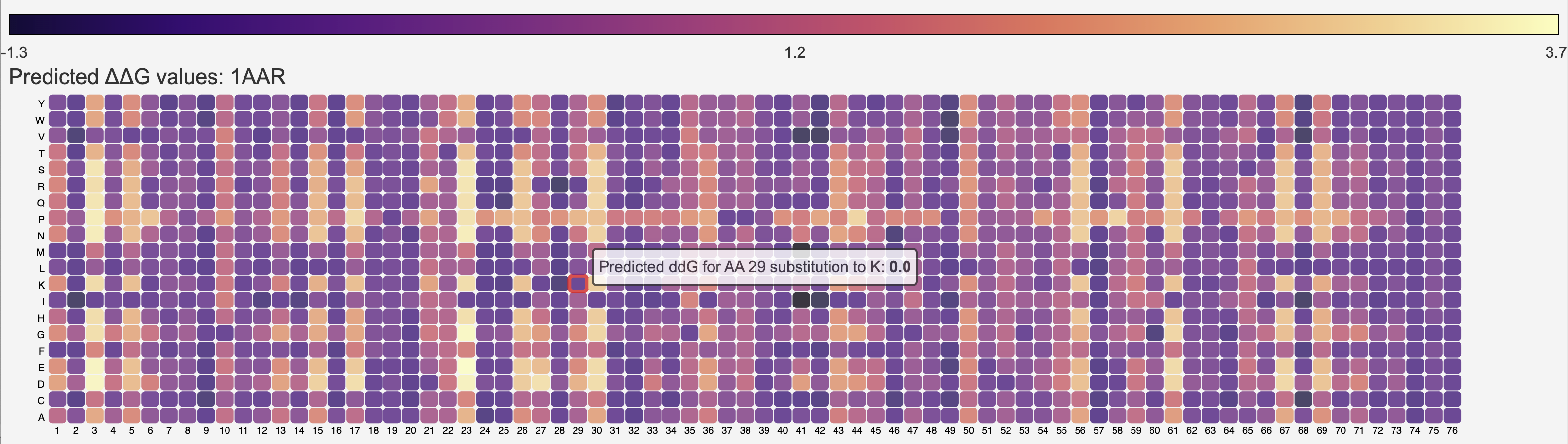}
\end{figure}

The deeper purple color indicates a more strongly negative $\Delta\Delta G$ while the lighter yellow color indicates a more strongly positive change. A single-point mutation that increases $\Delta\Delta G$ is preferred, as it makes a protein more thermostable.

\vspace{0.3cm} In summary, our approach introduces the following innovative techniques:
\begin{enumerate}
    \item Latent transfusion of embeddings leveraging several SOTA protein models, namely ESM and ThermoMPNN, in order to generate more accurate predictions of $\Delta\Delta G$ resulting from single-point protein mutations
    \item Intuitive and interactive web app detailing protein structure and thermostability changes due to single-point mutations meant to accelerate the existing drug discovery workflow
\end{enumerate}

Together, these advancements aim to make a meaningful impact in the realm of drug discovery.

\section{Experiments}

\hspace{0.25cm} The experiments in this study are designed to investigate if and which complementary information captured by different models, can help predict the impact of amino acid changes on protein stability with the highest accuracy. Additionally, our UI experiments aim to determine the most effective ways to present and visualize protein stability data, focusing on the optimal presentation of 3D structures for comparison. The ultimate goal is to create an intuitive, informative, and interactive interface that allows users to easily explore the impact of mutations on protein stability.

Table 2 presents the performance of our four models in predicting protein stability changes after conducting grid search and fine-tuning while also testing regularization techniques as needed. Model 3, the enhanced ThermoMPNN+ model, demonstrates the best performance among the four models, with the highest validation R², lowest MSE and RMSE, and the highest Spearman correlation coefficient. This suggests that the element-wise integration of these diverse feature sets and embeddings through latent transfusion captures important interactions between them, improving the model's predictive ability. The original ThermoMPNN model (Model 1) performs moderately well, while Model 2 and Model 4 show lower performance. Despite the rigorous optimization efforts, the overall performance of the models remains limited, with the best model explaining only 19.62\% of the variance in the $\Delta\Delta G$ values. This indicates that there is still room for improvement in predicting protein stability changes, and further research may be necessary to identify additional features, modeling techniques, or data sources to further enhance these models.

\begin{table}[!ht]
    \centering
    \caption{\textbf{Regression Evaluation Results}} \label{tab:title}
    \begin{tabular}{p{0.25\linewidth} p{0.15\linewidth} p{0.15\linewidth} p{0.15\linewidth}}
        \textbf{Model} & \textbf{MSE} & \textbf{R$^2$} & \textbf{Spearman} \\
        \hline
        Model 1 & 2.195 & 0.168 & 0.542 \\
        Model 2 & 2.397 & 0.064 & 0.387 \\
        \textbf{Model 3} & \textbf{2.057} & \textbf{0.196} & \textbf{0.552} \\
        Model 4 & 2.490 & 0.027 & 0.354 \\
        \bottomrule
    \end{tabular}
    \\
    \textit{Model 1 is the ThermoMPNN model. Model 3 is the ThermoMPNN+ model.}
\end{table}

Figure 5 shows t-SNE plots of processed combined embeddings from the ThermoMPNN+ model for the training (left) and validation (right) sets, with points representing proteins and colors indicating $\Delta\Delta G$ values or stability changes due to mutations. Interestingly, when visualizing the t-SNE embeddings from the ThermoMPNN+ model, we noticed distinct clusters forming across the range of $\Delta\Delta G$ values unlike the processed embeddings from our other models. Using the embeddings with a random forest classifier from PCA, another dimensionality reduction technique, we find better accuracy, precision, recall, and F1-score compared to the original ThermoMPNN model. These results suggest that the validation embeddings from the ThermoMPNN+ model are strong predictors of the stabilizing or destabilizing impact of mutations (or positive and negative changes in $\Delta\Delta G$, where positive values represent a destabilizing effect and negative values represent a stabilizing effect), resulting in a lower dimensional representation that enhances class separability. Future research could explore this approach, aligning with our study's goal of investigating complementary information captured by different models and techniques.


\begin{figure}[h]
    \centering
    \caption{\textbf{Data Visualization: Embedding Diversity in Protein Mutations}}
    \includegraphics[scale=0.25]{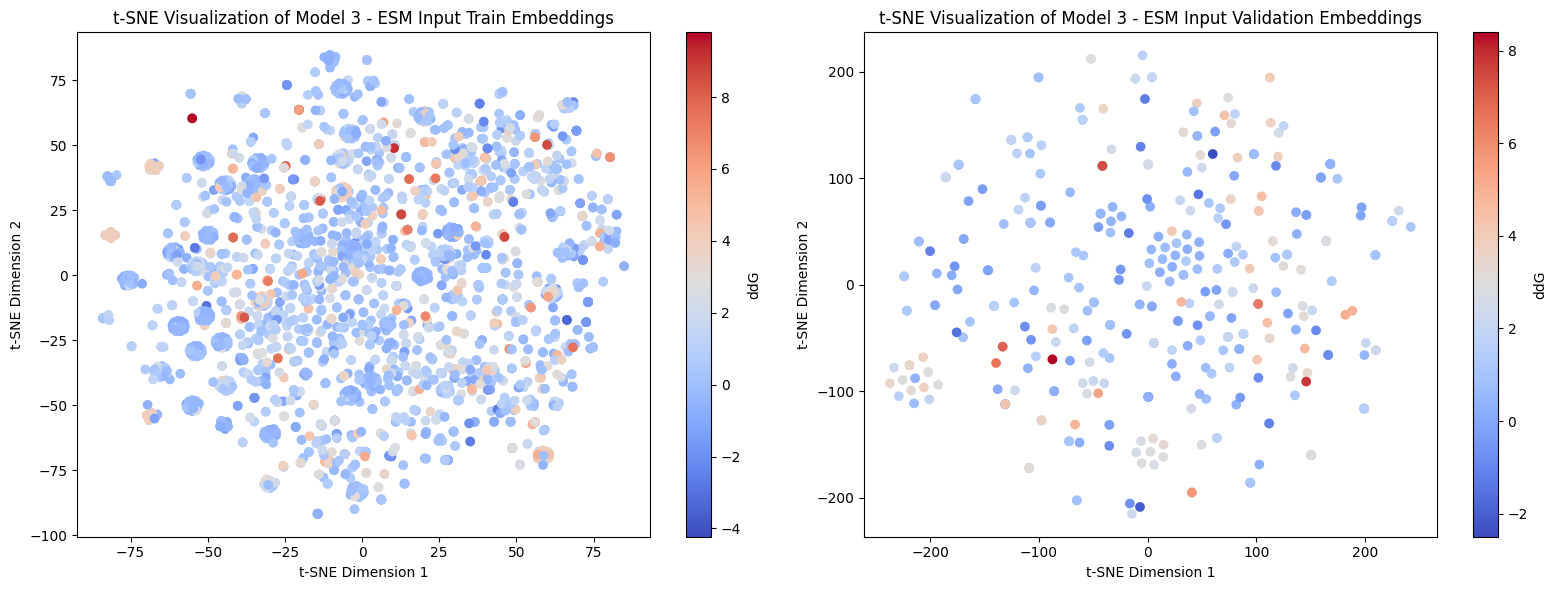}
\end{figure}


\begin{table}[!ht]
    \centering
    \caption{\textbf{Classification Results}} \label{tab:title}
    \begin{tabular}{p{0.35\linewidth} p{0.1\linewidth} p{0.1\linewidth} p{0.12\linewidth} p{0.1\linewidth}}
        \textbf{Model} & \textbf{Acc.} & \textbf{Prec.} & \textbf{Recall} & \textbf{F1} \\
        \hline
        ThermoMPNN & 0.770 & 0.876 & 0.889 & 0.882 \\
        ThermoMPNN+ & \textbf{0.859} & \textbf{0.859} & \textbf{1.000} & \textbf{0.924} \\
        \bottomrule
    \end{tabular}
\end{table}

We made several experimental revisions to our user interface to enhance user experience and accessibility. Initially, the 3D structures of the wild type and mutations were presented on separate pages, along with a static $\Delta\Delta G$ matrix, which was not user-friendly. We improved upon this by displaying both the wild type and mutation 3D structures on the same page in the next iteration. We also implemented an interactive $\Delta\Delta G$ matrix, significantly enhancing user interaction and making the information more accessible.




\section{Conclusions}
\hspace{0.25cm} We aimed to improve the drug discovery process by accurately predicting protein thermostability changes caused by single-point mutations. Our novel deep learning model, ThermoMPNN+, improves upon the state-of-the-art by incorporating features from a robust protein transformer model. We also developed an intuitive UI for users to interact with 3D protein structures and explore the impact of mutations on thermostability in real-time. These tools will significantly reduce the time and cost associated with evaluating proteins for new drugs, ultimately leading to faster and improved drug discovery outcomes.


During the modeling phase, we encountered several limitations. Despite attempting to augment the feature space with our various modeling techniques, the limited availability of data likely contributed to the still less than ideal ability of ThermoMPNN+, our best-performing model, to explain the variability in the data, as evidenced by the relatively low R² values. Model interpretability remains a challenge, as the underlying features used to predict $\Delta\Delta G$ are unknown. Computational complexity is also a significant limitation, with model training taking approximately 3+ hours using GPU. Despite these limitations, ThermoMPNN+ demonstrates improved performance compared to existing models, suggesting that integrating diverse feature sets and embeddings through latent transfusion techniques is a promising approach.

Future extensions of this research can include reframing the problem as a binary classification task, predicting whether mutations are stabilizing or destabilizing. Additionally, incorporating other types of information, such as B-factor (a measure of an atom's vibrational motion) and surface area information, could further enhance the model's predictive capabilities according to our domain research. The web app can be improved as well, allowing users to upload their own proteins to generate structural change and $\Delta\Delta G$ predictions on the fly. Additional UI improvements can focus on maximizing the user experience and utility of the application. As we refine these methods, considering trade-offs between model complexity, computational resources, and interpretability is crucial to develop robust and practical tools for protein engineering and drug discovery. 

All team members have contributed a similar amount of effort.

\newpage
\setcounter{footnote}{1}
\bibliography{mybib} 

\end{document}